\newcommand{\draft}{F}

\if T\draft
\documentclass[12pt,draft,twoside]{IEEEtran} % use for submission to IEEE
\else
\documentclass[twocolumn,twoside]{IEEEtran} % simulate final product
\fi

\usepackage{psfig}
\usepackage{amsmath}
\usepackage{amssymb}

\newcommand{\briefrefs}{T}

\if T\briefrefs

\else

\fi

\newcommand{\beqn}{\begin{equation}}
\newcommand{\eeqn}{\end{equation}}

\newcommand{\be}{\begin{equation}}
\newcommand{\ee}{\end{equation}}
\newcommand{\beqa}{\begin{eqnarray}}
\newcommand{\eeqa}{\end{eqnarray}}
\newenvironment{vectra}[1]%
	{\left[ 		
	\begin{array}{#1}}
	{\end{array} \right]}

\newcommand{\bv}{\begin{vectra}}
\newcommand{\ev}{\end{vectra}}

\newcommand{\ba}{\begin{array}}
\newcommand{\ea}{\end{array}}

\newcommand{\Comment}[1]{}

\newcommand{\figureLongCapTwoCol}[3]{
	\if F\draft
	\begin{figure}[htbp]
	\centerline{\psfig{figure=#2,width=3.25in}} % goes with \input psfig
	\caption{\protect {\small #3}}
	\label{#1} % LABEL MUST GO AFTER CAPTION OR INSIDE ITS ARGUMENT
	\end{figure}
	\fi
}

\begin{document}

\title{Orthogonal Least Squares Algorithm for the Approximation of a Map and its Derivatives with a RBF Network}
%\author{C. D.\thanks{C. D is...}}
\author{Carlo Drioli and Davide Rocchesso \thanks{Manuscript received \ldots}
\thanks{Carlo Drioli is with the Dipartimento di Elettronica e Informatica,
Universit\`a di Padova, 35131 Padova, Italy (e-mail: adrian@dei.unipd.it)}
\thanks{Davide Rocchesso is with the Dipartimento Scientifico e Tecnologico,
Universit\`a di Verona, 37134 Verona, Italy (e-mail: rocchesso@sci.univr.it)}
\thanks{This work has been submitted to the IEEE Transactions on Systems, Man, and Cybernetics -- part B, for possible publication. Copyright may be  transferred without notice, after which this version may no longer be accessible.}
}

\markboth{IEEE Transactions on Systems, Man, and Cybernetics-Part
B: Cybernetics, vol. XX, no. X}{Drioli and Rocchesso: Approximation of
a Map and its Derivatives with a RBF Network}

\maketitle

\begin{abstract}
Radial Basis Function Networks (RBFNs) are used primarily to
solve curve-fitting problems and for non-linear system modeling.
Several algorithms are known for the approximation of a non-linear
curve from a sparse data set by means of RBFNs. However, there are
no procedures that permit to define constrains on the
derivatives of the curve. In this paper, the Orthogonal Least
Squares algorithm for the identification of RBFNs is modified to
provide the approximation of a non-linear 1-in 1-out map along
with its derivatives, given a set of training data. The interest
on the derivatives of non-linear functions concerns many
identification and control tasks where the study of system stability and
robustness is addressed. The effectiveness of the
proposed algorithm is demonstrated by a study on the stability of
a single loop feedback system.
\end{abstract}


\section{Introduction}
\PARstart{T}{he} Orthogonal Least Squares (OLS) algorithm
\cite{ChenCowanGrant91} is one of the most efficient procedures
for the training of Radial Basis Function Networks (RBFN). A RBFN
is a two-layer neural network model especially suited for
non-linear function approximation, and appreciated in the fields
of signal processing \cite{NNHaykin94,HaykPrinc98}, non-linear
system modeling, identification and control
\cite{ChenBillings92,LiuKadirkBillings99,LangariWangYen97}, and
time-series prediction \cite{YeeHayk99,Casdagli89}.

Despite of the fact that in many identification and control tasks
the stability of the identified system depends on the gradient of
the map \cite{RomGrebOttDay92,ConnMartAtlas94}, the problem of
efficiently approximating a non-linear function along with its
derivatives seems to be rarely addressed. In
\cite{HornStinchWhite90,CardEuvrard92}, some theoretical results
as well as some application examples are found that apply to
generic feedforward neural networks.

In this paper, an extended version of the OLS algorithm for the
training of $1$-in $1$-out RBFNs is proposed, which permits to
approximate an unknown function by specifying a set of data points
along with its desired first-order derivatives.

The paper is organized as follows: in Section \ref{sec:OLS1}, the
OLS algorithm is reviewed and modified to add control over the
derivative of the function to be approximated. The extension to
higher order derivatives is introduced in Section \ref{sec:OLS2}.
Application examples in the field of single loop feedback systems
are given in Section \ref{sec:Apps}. In Section \ref{sec:concl},
the conclusions are presented.

\section{Orthogonal Least Squares Learning Algorithm}\label{sec:OLS1}

The OLS learning algorithm is traditionally tied to the parametric
identification of RBF networks, a special two-layer neural network
model widely used for the interpolation and modeling of data in
multidimensional space. In the following we will restrict the
discussion to the 1-in 1-out RBFN model, which is a mapping
$f:\;\mathbb{R} \rightarrow \mathbb{R}$ of the form
\begin{equation}\label{rbffun}
  f(x)=b+\sum_{i=1}^Hw_{i}\phi(x,m_{i}),
\end{equation}
where $x\in \mathbb{R}$ is the input variable, $\phi(\cdot)$ is a
given non-linear function, $b$, $w_i$ and $m_i$, $1 \leq i\leq H$,
are the parameters, and $H$ is the number of radial units. The
RBFN can be viewed as a special case of the linear regression
model
\begin{equation}\label{regression}
  t(k)=b+\sum_{i=1}^Hw_{i}p_i(k)+e(k),
\end{equation}
where $t(k)$ is the desired $k$-th output sample, $e(k)$ is the approximation
error, and $p_i(k)$ are the regressors, i.e. some fixed functions
of $x(k)$, where $x(k)$ are the input values corresponding to the
desired output values $t(k)$:
\begin{equation}
p_i(k)=\phi(x(k),m_i).
\end{equation}

%Next, we will briefly recall the classic OLS algorithm and
%illustrate the modifications to achieve the control on
%derivatives.

In its original version, the OLS algorithm is a procedure
 iteratively selects the best regressors (radial basis units)
from a set of available regressors. This set is composed of a
number of regressors equal to the number of available data, and
each regressor is a radial unit centered on a data point. The
selection of radial unit centers is recognized as the main problem
in the parametric identification of these models, while the choice
of the non-linear function for the radial units does not seem to
be critical. Although gaussian-shaped functions are often
preferred, spline, multi-quadratic and cubic functions are valid
alternatives. Here, we will use the cubic function $\phi
(x,m)=(\|x-m\|)^3$, where $\| \cdot\|$ denotes the euclidean norm
and $m$ denotes the center of the radial unit.

\subsection{Classic OLS algorithm}

Say $\{x(k),t(k)\}$, $k=1,2,...,N$, is the data set given by $N$
input-output data pairs, which can be organized in two column
vectors ${\mathbf x}=[x(1)\cdots x(N)]^T$ and ${\mathbf
t}=[t(1)\cdots t(N)]^T$. The model parameters are given in vectors
${\mathbf m}=[m_{1}\cdots m_{H}]^T$, ${\mathbf w}=[w_{1}\cdots
w_{H}]^T$ and ${\mathbf b}=[b]$, where $H$ is the number of radial
units to be used. Arranging the problem in matrix form we have:
\begin{equation}\label{vecols0}
    {\mathbf t}=\left[\begin{array}{cc}
    {\mathbf P} & {\mathbf 1} \end{array}\right]
    \left[ \begin{array}{c}
      {\mathbf w} \\ b \end{array}\right]+ {\mathbf e}
\end{equation}
with
\begin{eqnarray}
   \nonumber {\mathbf P} & = & \left[ \begin{array}{ccc}
       {\mathbf p}_1 & \cdots & {\mathbf p}_H \end{array}\right] \\
       & = & \left[ \begin{array}{ccc}
      \phi(x(1),m_{1}) & \cdots & \phi(x(1),m_{H})\\
      \vdots & \ddots & \vdots \\
      \phi(x(N),m_{1}) & \cdots & \phi(x(N),m_{H})
    \end{array}\right],
\end{eqnarray}
where ${\mathbf p}_i=[\phi (x(1),m_i)\dots \phi (x(N),m_i)]^T$ are
regressor vectors forming a set of basis vectors, ${\mathbf
e}=[e(1)\cdots e(N)]^T$ is the identification error, and ${\mathbf
1}=[1\dots 1]^T$ is a unit column vector of length $N$. The least
squares solution of this problem satisfies the condition that
\begin{equation}
\tilde{{\mathbf t}}=\left[\begin{array}{cc}
    {\mathbf P} & {\mathbf 1} \
    \end{array}\right]\left[ \begin{array}{c}
      {\mathbf w} \\ b
    \end{array}\right]
\end{equation}
is the projection of ${\mathbf t}$ in a vector space spanned by
the regressors. If the regressors are not independent, the
contribution of each regressor to the total energy of the desired
output vector is not clear. The OLS algorithm iteratively selects
the best regressors from a set by applying a Gram-Schmidt
orthogonalization, so that the contribution of each vector of this
new orthogonal base can be determined individually among the
available regressors.

\subsection{Modified OLS algorithm}
The classic algorithm selects the best set of regressors from the
ones available, and determines the output layer weights for the
identification of the desired in-out map, but does not explicitly
controls the derivative of the function. We propose to modify this
procedure so to permit to specify the desired value of the
function derivative in each data point. The data set will then be
organized in three vectors ${\mathbf x}=[x(1)\cdots x(N)]^T$,
${\mathbf t}=[t(1)\cdots t(N)]^T$, and ${\mathbf
t}^{(1)}=[t_1(1)\cdots t_1(N)]^T$, ${\mathbf x}$ and ${\mathbf t}$
being the input-output pairs and ${\mathbf t}^{(1)}$ being the
respective derivatives. It has to be noted that the original OLS
algorithm selects each radial unit from a set of units, each of
which is centered on a input data point. The maximum number of
units is then limited to the number of data points. When we add
requirements on the derivative of the function, a further
constraint to the optimization problem is added, and the number of
units to be selected in order to reach the desired approximation
may be higher then the number of data points. A possible choice is
to augment the input vector with points where there is no data
available, and to build the set of $N_e$ regressors on this
extended vector. 

The algorithm can be summarized as follows:
\begin{itemize}
  \item {\bf First step}, initialization: the set of regressors for selection is obtained by centering
  the $N_e$ radial units, and the {\em error reduction ratio} (err) for
  each regressor vector is computed. Given the regressor vectors
\begin{equation}
  {\mathbf p}_i=[\phi(x(1),x(i)),...,\phi(x(N),x(i))]^T,\;\; 1\leq
i\leq N_e,
\end{equation}
and defined the first-iteration vectors
\begin{equation}
  {\mathbf u}_{1,i}={\mathbf p}_i,\;\; 1\leq i\leq N_e.
\end{equation}
The error reduction ratio associated with the $i$-th vector is
given by
\begin{equation}
\mbox{err}_{1,i}=({\mathbf u}_{1,i}^T{\mathbf t})^2/(({\mathbf
u}_{1,i}^T {\mathbf u}_{1,i})({\mathbf t}^T{\mathbf t})).
\end{equation}
In a similar way, the regressor vectors for the derivative of the
map are computed:
\begin{equation}
{\mathbf p}_i^{(1)}=[\frac{\partial\phi(x(1),x(i))}{\partial
x}\cdots \frac{\partial\phi(x(N),x(i))}{\partial x}]^T,\;\; 1\leq
i\leq N_e,
\end{equation}
and the first-iteration vectors are defined:
\begin{equation}
 {\mathbf l}_{1,i}={\mathbf p}_i^{(1)},\;\; 1\leq i\leq
  N_e.
\end{equation}
The error reduction ratio for the derivative is:
\if T\draft
\begin{equation}
\mbox{grad\_err}_{1,i}=({\mathbf l}_{1,i}^T {\mathbf
t}^{(1)})^2/(({\mathbf l}_{1,i}^T{\mathbf l}_{1,i})({{\mathbf
t}^{(1)}}^T{\mathbf t}^{(1)})), 1\leq i\leq N_e.
\end{equation}
\else
\begin{eqnarray}
\mbox{grad\_err}_{1,i}=({\mathbf l}_{1,i}^T {\mathbf
t}^{(1)})^2/(({\mathbf l}_{1,i}^T{\mathbf l}_{1,i})({{\mathbf
t}^{(1)}}^T{\mathbf t}^{(1)})),\\ \nonumber  1\leq i\leq N_e.
\end{eqnarray}
\fi
The $\mbox{ err}_{1,i}$ and $\mbox{ grad\_err}_{1,i}$ represent
the error reduction ratios caused respectively by ${\mathbf
u}_{1,i}$ and ${\mathbf l}_{1,i}$, and the total error reduction
ratio can be computed by
\begin{equation}
\mbox{ tot\_err}_{1,i}=\lambda \mbox{ err}_{1,i}+(1-\lambda)\mbox{
grad\_err}_{1,i},
\end{equation}
where $\lambda$ weights the importance of the
map against its derivative. The index $i_1$ is then found, so
that:
\begin{equation}
\mbox{tot\_err}_{1,i_1}=\max_i \{\mbox{tot\_err}_{1,i},\;\; 1\leq
i\leq N_e\}.
\end{equation}

The regressor ${\mathbf p}_{i_1}$ giving the largest error
reduction ratio is selected and removed from the set of available
regressors. The corresponding center is added to the set of
selected centers:
\begin{equation}
{\mathbf u}_1={\mathbf u}_{1,i_1}={\mathbf p}_{i_1};
\end{equation}
\begin{equation}
{\mathbf l}_1={\mathbf l}_{1,i_1}={\mathbf p}_{i_1}^{(1)};
\end{equation}
\begin{equation}
m_1=x(i_1).
\end{equation}

  \item {\bf $h$-th iteration}, for $h=1,...,H$ and $H\leq N_e$:
  the regressors selected in the previous steps, having indexes $i_1,...,i_{h-1}$,
  have been removed from the set of available regressors. Before computing
  the error reduction ratio for each regressor still available, the orthogonalization
  step is performed which makes each regressor orthogonal with
  respect to those already selected:

\begin{equation}
{\mathbf u}_{h,i}={\mathbf p}_i-\sum_{j=1}^{h-1}({\mathbf
u}_j^T{\mathbf p}_i)/({\mathbf u}_j^T{\mathbf u}_j){\mathbf
u}_j,\;\; i\neq i_1,i_2,...,i_{h-1};
\label{uhi}
\end{equation}

\begin{equation}
{\mathbf l}_{h,i}={\mathbf p}_i^{(1)}-\sum_{j=1}^{h-1}({\mathbf
l}_j^T{\mathbf p}_i^{(1)})/({\mathbf l}_j^T{\mathbf l}_j){\mathbf
l}_j,\;\; i\neq i_1,i_2,...,i_{h-1};
\end{equation}

\begin{equation}
\mbox{err}_{h,i}=({\mathbf u}_{h,i}^T{\mathbf t})^2/(({\mathbf
u}_{h,i}^T{\mathbf u}_{h,i})({\mathbf t}^T{\mathbf t})),\;\; i\neq
i_1,i_2,...,i_{h-1};
\end{equation}

\begin{eqnarray}
\mbox{ grad\_err}_{h,i}=({\mathbf l}_{h,i}^T {\mathbf
t}^{(1)})^2/(({\mathbf l}_{h,i}^T{\mathbf l}_{h,i}){({\mathbf
t}^{(1)}}^T{\mathbf t}^{(1)})), 
\label{graderrhi}
\\ \nonumber i\neq
i_1,i_2,...,i_{h-1};
\end{eqnarray}

\begin{eqnarray}
\mbox{ tot\_err}_{h,i}=\lambda \mbox{err}_{h,i}+(1-\lambda)\mbox{
grad\_err}_{h,i}, \\ \nonumber i\neq i_1,i_2,...,i_{h-1}.
\end{eqnarray}

As before, the regressor with maximum error reduction ratio is
selected and removed from the list of availability, and its center
is added to the set of selected centers:
\begin{equation}
\mbox{ tot\_err}_{h,i_h}=\max_i\{\mbox{ tot\_err}_{h,i},\;\; i\neq
i_1,i_2,...,i_{h-1} \}
\end{equation}
\begin{equation}
{\mathbf u}_h={\mathbf u}_{h,i_h};
\end{equation}
\begin{equation}
{\mathbf l}_h={\mathbf l}_{h,i_h};
\end{equation}
\begin{equation}
m_{h}=x(i_h).
\end{equation}

  \item {\bf Final step}, computation of output layer weights : once the $H$ radial units have been positioned,
  the remaining ${\mathbf w}$ and ${\mathbf b}$ parameters can be found with a Moore-Penrose matrix
  inversion: let us call ${\mathbf P}_H=[{\mathbf u}_1 {\mathbf u}_2 \cdots {\mathbf
  u}_H]$ and ${\mathbf P}_H^{(1)}=[{\mathbf l}_1 {\mathbf l}_2 \cdots {\mathbf
  l}_H]$ the two sets of selected regressors, and let ${\mathbf 1}=[1\dots 1]^T$ and
${\mathbf 0}=[0\dots 0]^T$ be two column vectors of length $N$. Then we
have
\begin{equation}\label{vecols}
   \left[ \begin{array}{c}
    {\mathbf t} \\
    {\mathbf t}^{(1)}\end{array}\right] =
   \left[ \begin{array}{cc}
      {\mathbf P}_H & {\mathbf 1} \\
      {\mathbf P}_H^{(1)} & {\mathbf 0}
    \end{array}\right]
    \left[\begin{array}{c}
      {\mathbf w} \\ {b} \end{array} \right]
      +{\mathbf e}_H
\end{equation}
whose solution is
\begin{equation}\label{vecols2}
 \left[\begin{array}{c}
 \mathbf{w} \\ {b} \end{array} \right]
   = \left(
   \left[ \begin{array}{cc}
      {\mathbf P}_H & {\mathbf 1} \\
      {\mathbf P}_H^{(1)} & {\mathbf 0}
    \end{array}\right] \right)^+
    \left[ \begin{array}{c}
    \mathbf{t} \\ {\mathbf t^{(1)}}
   \end{array}\right]
.\end{equation}
\end{itemize}

Usually, it is convenient to stop the procedure before the maximum
number of radial units has been reached, as soon as the identification
error is considered to be acceptable. To this purpose, one can use
equation (\ref{vecols2}) at iteration $h$ to compute the
identification error ${\mathbf e}_h$ in (\ref{vecols})
\footnote{Note that in this case the length of vector ${\mathbf
w}$ and the number of columns of matrices ${\mathbf P}$ in equation
(\ref{vecols2}) is $h$ instead of $H$}.

\subsection{Example}
Let us consider, as an example, the fitting of a step-like data set,
where the derivative is arbitrarily constrained. The data set is
shown in Fig. \ref{step1}, along with the result of the parametric
identification routine. It can be seen how an unlikely derivative
was chosen in the critical zone to highlight the properties of the
model. Fig. \ref{step2} shows the interpolating properties of the
resulting RBF Network when computed on an input interval which is
denser than the original input data set.

\if F\draft
\begin{figure}[h]
  \begin{center}
%\psfull
\psfig{file=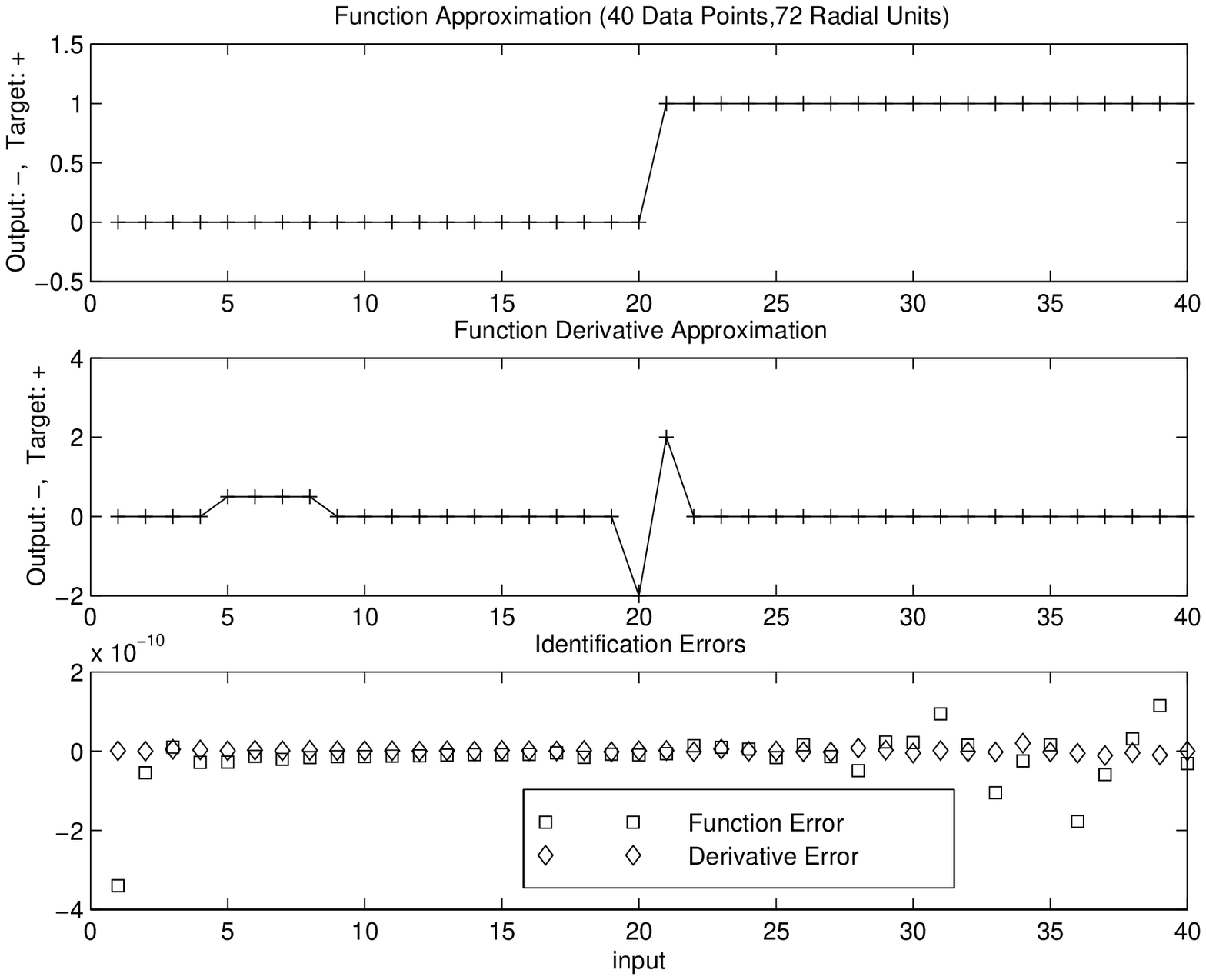,width=9cm}
\caption{Result of the training
  procedure applied to the fitting of a step-like data set ($+$, upper figure),
  with arbitrary derivative constraint ($+$, middle figure). In upper and middle figures,
  $+$ is the desired output and the continuous line is the actual output.
  The problem required $72$ radial units to fit $40$ data points, with an
  identification error less than $10^{-9}$ in magnitude.}\label{step1}
  \end{center}
\end{figure}

\begin{figure}[h]
  \centering
%\psfull
\psfig{file=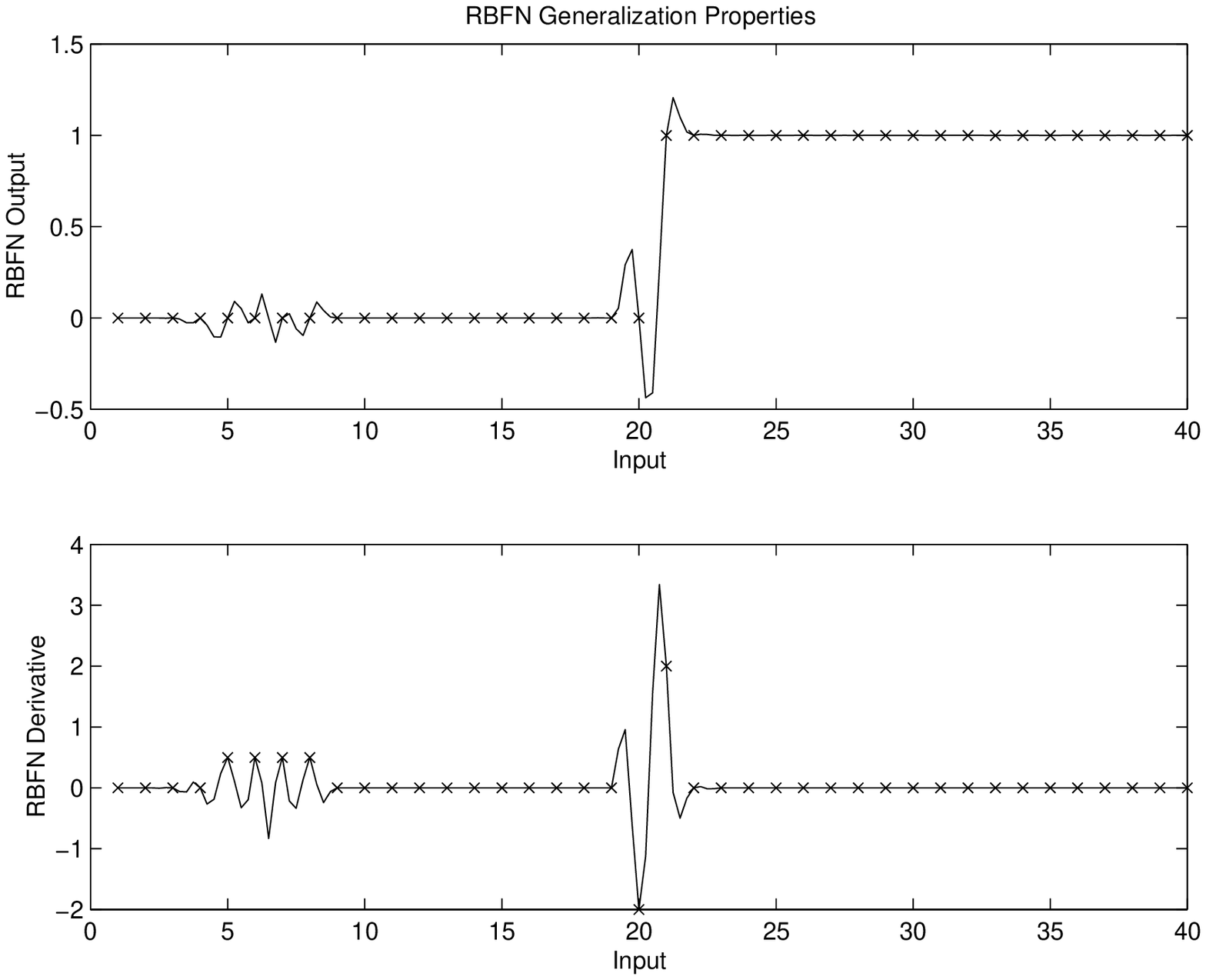,width=9.0cm}
  \caption{Interpolation properties of the
  identified RBF Network: the output of the model was computed on a input set which is
  denser than the original one (x: data set).}\label{step2}
\end{figure}
\fi
\section{Higher order derivatives}\label{sec:OLS2}
The extension of the algorithm for the identification of a map and
its derivatives of order higher than one is straightforward. Given
that $\phi$ is continuous and has continuous derivatives up to
order $r$, then
%$\phi\in{\mathcal C}^{(r)}$,
the derivatives of order up to $r$ can be identified for the map
$f$. The data set is organized in $r+1$ vectors ${\mathbf
x}=[x(1)\cdots x(N)]^T$, ${\mathbf t}=[t(1)\cdots t(N)]^T$,
${\mathbf t}^{(1)}=[t_1(1)\cdots t_1(N)]^T$, ..., ${\mathbf
t}^{(r)}=[t_r(1)\cdots t_r(N)]^T$, where $t^{(i)}(k)$ is the
desired $i$-th derivative for the $k$-th data point. In the first
step, a different set of regressors can be computed for each
derivative order:
\begin{equation}
 {\mathbf p}_i=[\phi(x(1),x(i)),\dots,\phi(x(N),x(i))]^T,\;\; 1\leq i\leq
 N_e;
\end{equation}
\begin{eqnarray}
 \nonumber
 {\mathbf p}_i^{(d)}=[\frac{\partial^d\phi(x(1),x(i))}{\partial x^d},\cdots,\frac{\partial^d\phi(x(N),x(i))}{\partial
 x^d}]^T, \;\; \\
 1\leq i\leq N_e,\;\; 1\leq d\leq r.
\end{eqnarray}
If we now call ${\mathbf u}_{i_{h-1}}$, ${\mathbf
l}^{(1)}_{i_{h-1}}$,$\cdots$, ${\mathbf l}^{(r)}_{i_{h-1}}$ the
orthogonalized regressor vectors selected in the $(h-1)$-th
iteration, in the $h$-th iteration the corresponding $r+1$ error
reduction ratios can be computed similarly to what shown
in equations~(\ref{uhi}--\ref{graderrhi}), and the total error reduction ratio can then be
computed as the weighted sum of these terms:

\begin{equation}
{\mathbf u}_{h,i}={\mathbf p}_i-\sum_{j=1}^{h-1}({\mathbf
u}_j^T{\mathbf p}_i)/({\mathbf u}_j^T{\mathbf u}_j){\mathbf
u}_j,\;\; i\neq i_1,i_2,...,i_{h-1};
\end{equation}

\begin{equation}
\mbox{err}_{h,i}=({\mathbf u}_{h,i}^T{\mathbf t})^2/(({\mathbf
u}_{h,i}^T {\mathbf u}_{h,i})({\mathbf t}^T{\mathbf t})),\;\;
i\neq i_1,i_2,...,i_{h-1};
\end{equation}

\begin{equation}
{\mathbf l}_{h,i}^{(d)}={\mathbf
p}_i^{(d)}-\sum_{j=1}^{h-1}({\mathbf l}_j^T{\mathbf
p}_i^{(d)})/({\mathbf l}_j^T{\mathbf l}_j){\mathbf l}_j,\;\;
i\neq i_1,i_2,...,i_{h-1};
\end{equation}

\begin{eqnarray}
\nonumber \mbox{err}_{h,i}^{(d)}=({{\mathbf l}_{h,i}^{(d)}}^T
{\mathbf t}^{(d)})^2/(({{\mathbf l}_{h,i}^{(d)}}^T{\mathbf
l}_{h,i}^{(d)})({{\mathbf t}^{(d)}}^T{\mathbf t}^{(d)})), \;\; \\
i\neq i_1,i_2,...,i_{h-1};
\end{eqnarray}
\begin{eqnarray}
\mbox{ tot\_err}_{h,i}=\lambda_0 \mbox{err}_{h,i}+\sum_{d=1}^r
\lambda_d \mbox{err}_{h,i}^{(d)}  \\ i\neq i_1,i_2,...,i_{h-1}.
\end{eqnarray}
The regressors with maximum error reduction ratio are selected and
removed from the list of availability, and the corresponding
centers are added to the set of selected centers:

\begin{equation}
\mbox{ tot\_err}_{h,i_h}=\max_i\{\mbox{ tot\_err}_{h,i},\;\; i\neq
i_1,i_2,...,i_{h-1} \};
\end{equation}

\begin{equation}
{\mathbf u}_h={\mathbf u}_{h,i_h};
\end{equation}

\begin{equation}
{\mathbf l}_h^{(d)}={\mathbf l}_{h,i_h}^{(d)},\;\; 1\leqslant d
\leqslant r;
\end{equation}

\begin{equation}
m_{h}=x(i_h).
\end{equation}

If we now let
\begin{equation}
\left\{
\begin{array}{ccc}
 {\mathbf P}_H & = & [{\mathbf u}_1\cdots {\mathbf
 u}_{H}]\\
 {\mathbf P}^{(1)}_H & = & [{\mathbf l}_1^{(1)}\cdots {\mathbf
 l}_{H}^{(1)}]\\
 &\vdots& \\
{\mathbf P}^{(r)}_H & = & [{\mathbf l}_1^{(r)}\cdots {\mathbf
 l}_{H}^{(r)}]
\end{array}
\right .
\end{equation}
be the final set of orthogonal regressors obtained from the
selection procedure, we can compute the output layer parameters by
solving the matrix equation
%\begin{table*}
%\centering
%\begin{equation}\label{hierinv}
%\left[ \begin{array}{cccc}
%    {\mathbf t} & {\mathbf t}^{(1)} & \cdots & {\mathbf t}^{(r)} \end{array}\right] =
%   \left[\begin{array}{cc}
%      {\mathbf w} & b \end{array} \right]
%   \left[ \begin{array}{cccc}
%      {\mathbf P}_H^T & ({\mathbf P}^{(1)}_H)^T & \cdots & ({\mathbf
%      P}^{(r)}_H)^T \\
%      {\mathbf 1}& {\mathbf 0} & \cdots & {\mathbf 0}
%    \end{array}\right]+{\mathbf e}.
%\end{equation}
%\end{table*}
\begin{equation}\label{hierinv}
\left[ \begin{array}{c}
    {\mathbf t} \\ {\mathbf t}^{(1)} \\ \vdots \\ {\mathbf t}^{(r)}
    \end{array}\right]=
   \left[ \begin{array}{cc}
      {\mathbf P}_H & {\mathbf 1} \\
      {\mathbf P}^{(1)}_H & {\mathbf 0} \\
      \vdots & \vdots \\
      {\mathbf P}^{(r)}_H & {\mathbf 0} \\
    \end{array}\right]
    \left[\begin{array}{c}
      {\mathbf w} \\ b \end{array} \right]+{\mathbf e}.
\end{equation}

\section{Application Examples}\label{sec:Apps}
The OLS algorithm for the identification of a map and its
derivatives with RBF networks is demonstrated using some examples
from the field of feedback non-linear systems.

\subsection{Single loop feedback system and the Hopf bifurcation
Theorem}\label{sec:Hopf} The single loop feedback circuit depicted
in Fig. \ref{feedback} is an example of autonomous non-linear
system capable of different dynamical behaviors, such as decaying
oscillation, stable periodic motion (including constant), and
chaos.
\if F\draft
\begin{figure}[h]
  \centering
%\psfull
\psfig{file=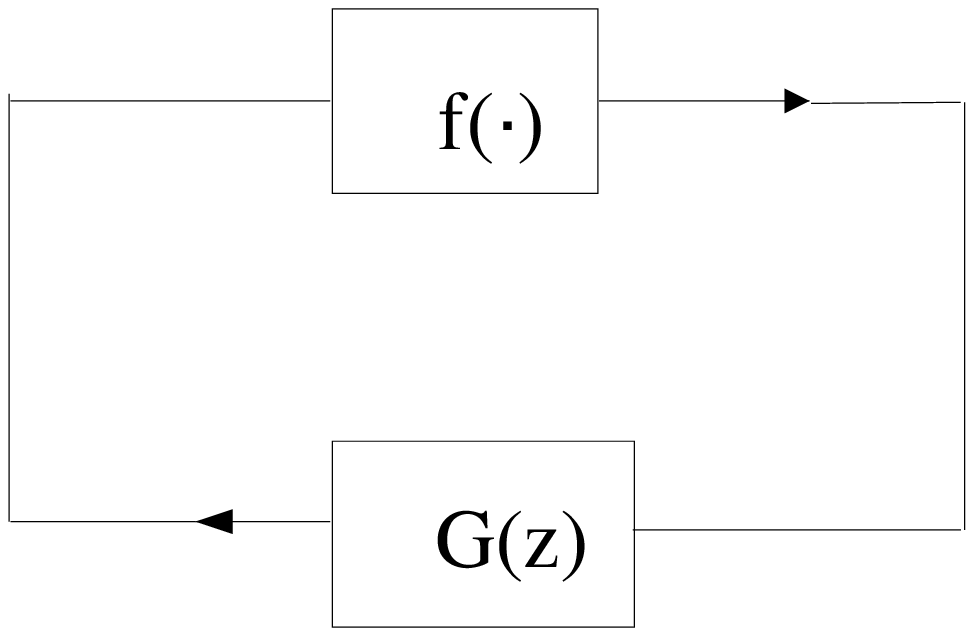,width=9.0cm}
\caption{Single loop feedback system}
\label{feedback}
\end{figure}
\fi
We will consider the case where $G(z)$ is made of two cascaded
linear elements, i.e. a delay line $D_L(z)$ of given length $L$,
and a low-pass filter $H(z)$. The function $f$ is assumed to be a
three-fixed points smooth function crossing the origin with slope
$S_1$, and having slopes $S_2$ and $S_3=S_2$ in the other two
points (see Fig. \ref{chuatest1}-a). The topology of fig.~\ref{feedback} is of
particular interest in the field of sound synthesis, for the
physically inspired modeling of musical instruments with sustained
sound \cite{McIntyreSchumWood83,Rodet93}, and has been object of
investigation by the authors for the construction of
generalized musical tone generators \cite{DrioRoc98}. The length
of the delay line, which can be seen as the medium where sound
propagates (such as a flute pipe or a violin string), is inversely
proportional to the pitch of the signal generated, and represents
an example of sound control parameter with a clear physical
meaning. The shape of the non-linear map and its fixed-point
derivatives are recognized to be responsible for the stability of
periodic motion, for the spectral content of the signal, and for
the time-constant of transient extinction. We don't care here
about the shape of the map, and we focus on the fixed points and
their derivatives. The condition for instability of the fixed
point in the origin, and thus the condition for the system to
oscillate, can be stated in terms of the Nyquist plot of the open
loop transfer function $G(z)=D_L(z)H(z)$. Say that $-q+j0$ denotes
the leftmost intersection point of the Nyquist plot of $G(z)$ with
the real axis. In order to let the system oscillate, a necessary
condition for $S_1$ is $S_1<-1/q$ \cite{Rodet93}. A different role
is assumed for the slope $S_2$, which is responsible for limiting
the growth of the system state. To this purpose, a slope $S_2>-1$
is needed at some distance from the origin, in correspondence of
the other two fixed points.

As a practical example, a low-pass filter $H(z)=0.4+0.3z^{-1}$ and
a delay length $L~=~100 \; \mbox{samples}$ are considered. The Nyquist plot of $G(z)$
has the smallest intersection point with the real axis in
$-0.7+j0$ which gives a maximum slope of -1.4286 over which the
oscillation will not occur. The length $L=100$ for the delay line
gives a period length $T_p=200$, which corresponds to a pitch of
$110.25$ Hz at a sample rate of $22050$ Hz. In Fig.
\ref{chuatest1}, the time evolution of the system is shown for a
map $f$ with fixed points $(0,0)$, $(-100,100)$, $(100,-100)$, for
different values of the slope $S_1$, and for a random initial
state in the range $[-0.1,0.1]$. It can be seen that the slope
$S_1$ can be used to drive the system to a periodic steady state
and to control the transient velocity.
%Note sullo spazio $(S_1,S_2)$, regioni stabili, regioni caotiche.
\if F\draft
\begin{figure}[h]
  \centering
%\psfull
\psfig{file=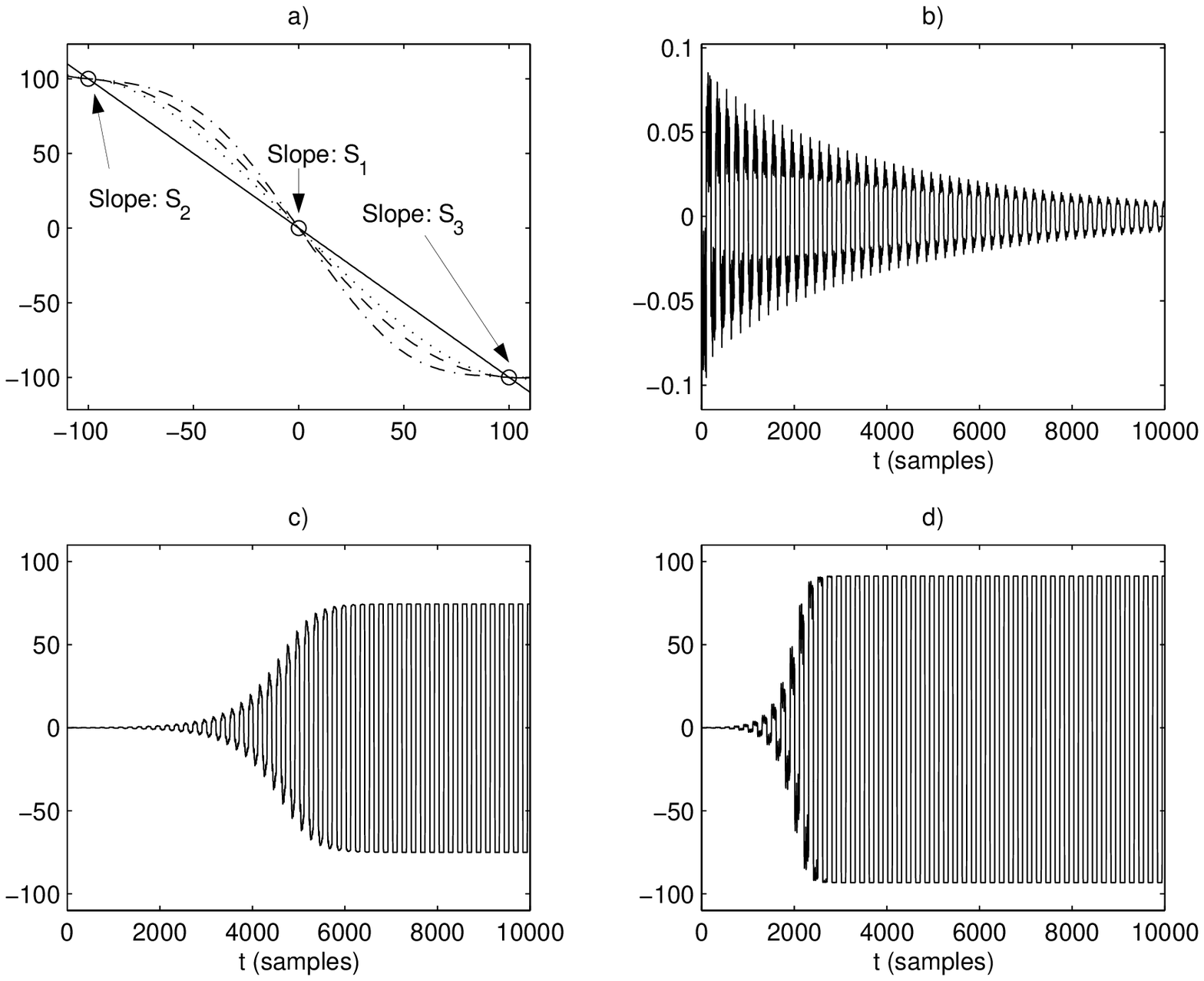,width=9.0cm}
\caption{Simulation of the circuit for different
  values of the derivative of the fixed point in the origin. a) shapes of the function $f(\cdot)$ for
  $S_1=-1.4$ (dotted line),  $S_1=-1.65$ (dashed line) and  $S_1=-2$ (dashdotted
  line). b,c,d) time evolution from random initial conditions in the three cases. The slope
  $S_2=-0.1$ of the other two fixed points is held constant in the three cases,
  and limits the growth of the system evolution.}\label{chuatest1}
\end{figure}
%\begin{figure}[h]
%  \centering
%  \includegraphics[scale=.5]{chuatest1fft.eps} \caption{Frequency analysis of a 1024 points windowed
%  frame around the $6000$-th sample of each output time series}\label{chuatest1fft}
%\end{figure}
\fi

The example shown is a particular case of the more general Hopf
bifurcation theorem \cite{Chua84} in its frequency domain
formulation. It is interesting to point out that the single loop
feedback systems exemplified in \cite{Chua84} are discretized
versions of simple $RLC$ electrical circuits, with at least a
non-linear component (e.g., a tunnel diode). The result is a
feedback scheme as the one in Fig. \ref{feedback}, where $G(z)$ is
a second-order $IIR$ transfer function, and no delay lines are
considered in the loop. In these circuits, a stable almost
sinusoidal oscillation is reached, whose frequency and amplitude
are functions of the second and third derivatives of the
non-linear map $f$, evaluated in the equilibrium point (i.e., dc
operating point), which is solution of the equation
$G(0)f(y)-y=0$.

\subsection{Stability control in feedback systems}\label{sec:stab}
Still referring to the closed loop feedback system of Fig.
\ref{feedback}, we are now interested in the stabilization of a
given periodic motion. With respect to the case of Section
\ref{sec:Hopf}, we're facing the dual situation, where we ignore
the transient part of the process and we're interested in the
shape of the period of the resulting time series.

Let us call ${\mathbf y}=[y_1,y_2,\dots,y_{T_p}]^T$ the desired
period and assume that the length of the period is even, i.e.
$T_p=2L$. For simplicity we consider the case that the filter
$H(z)$ is not present, thus the linear system $G(z)$ is  just
a delay line $D_L(z)$, which has to be of length $L$, as seen in
the previous example. The construction of the non-linear map able
to produce the desired periodic waveform is straightforward, and
relies on the training set $\mathcal{Y}$ computed using the data
points:
\begin{equation}\label{eqn:cup}
{\mathcal Y}={\mathcal Y}_1 \cup {\mathcal Y}_2,
%= \left\{
%(y_{i},y_{L+i}) \right\} \cup \left\{
%    (y_{L+i},y_i),
%\right\}
\end{equation}
where
\begin{equation}
{\mathcal Y}_1=\bigcup_{k=1}^{L}(y_k,y_{L+k}),
\end{equation}
and
\begin{equation}
{\mathcal Y}_2=\bigcup_{k=1}^{L}(y_{L+k},y_k).
\end{equation}

In Fig. \ref{halfper2id}, the computation of the training set from
the desired output process is illustrated, as well as the
approximation of the unknown function given by the proposed
algorithm.
\if F\draft
\begin{figure}[h]
  \centering
%\psfull
\psfig{file=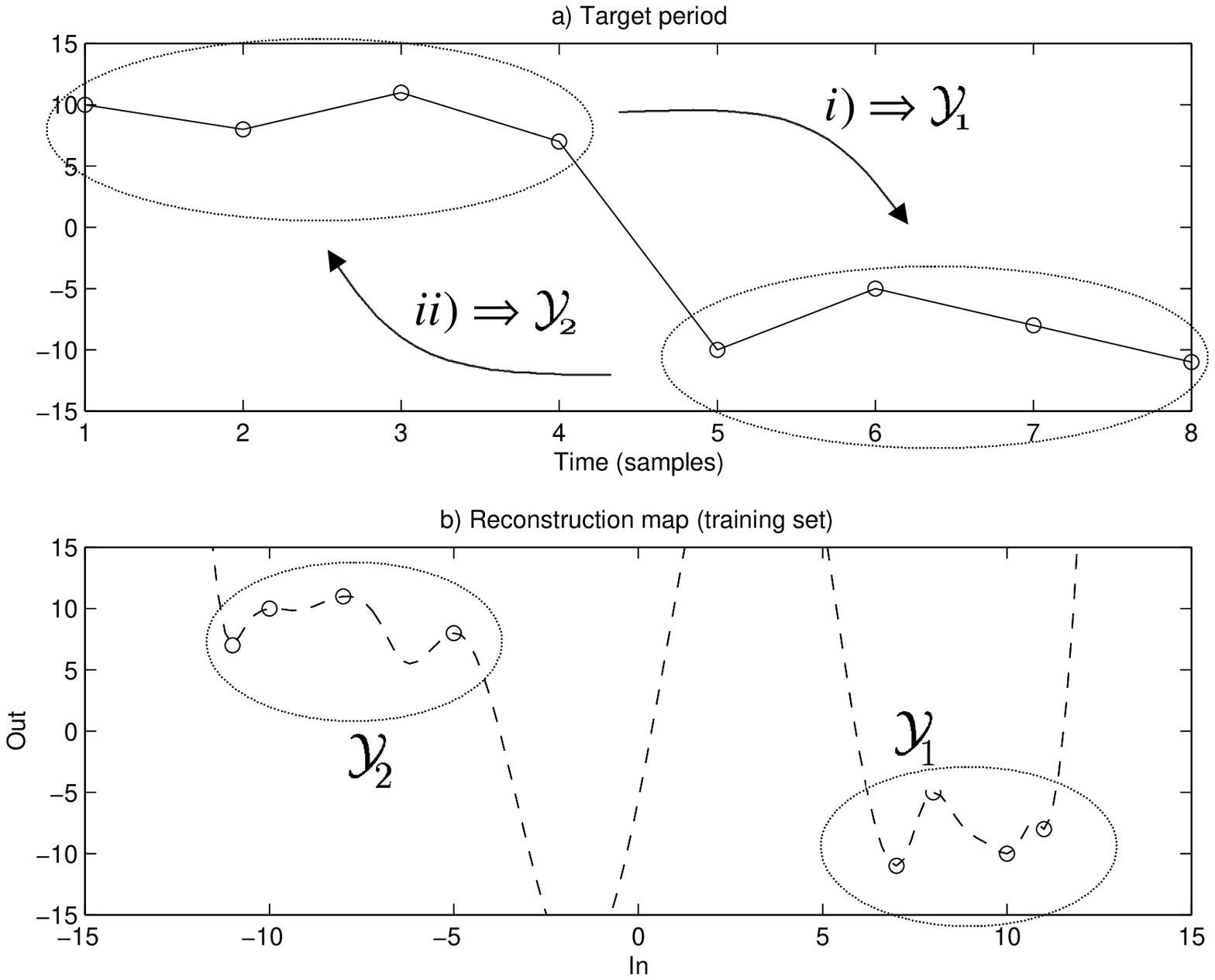,width=8.0cm}
  \caption{Training data (a) and approximation of the unknown
  function $f$ (b, dashed curve). A desired derivative of $0.3$ in
magnitude was imposed for all eight data points}
  \label{halfper2id}
\end{figure}
\fi
If the system state is initialized with a half-period, i.e. is
${\mathbf x_0}=[y_1,y_2,\dots,y_{L}]^T$, the non-linear map
iteratively computes the other half. The stability and robustness
with respect to additive noise is granted by the derivative of the
map, which has to be less then one in magnitude. Fig.
\ref{halfpersimu} shows the time evolution of the system whose
non-linear map data point derivatives are constrained to a
magnitude of $0.3$, and whose evolution is temporarily disturbed
with additive noise, with a SNR of $46$ dB.
\if F\draft
\begin{figure}[h]
  \centering
%\psfull
\psfig{file=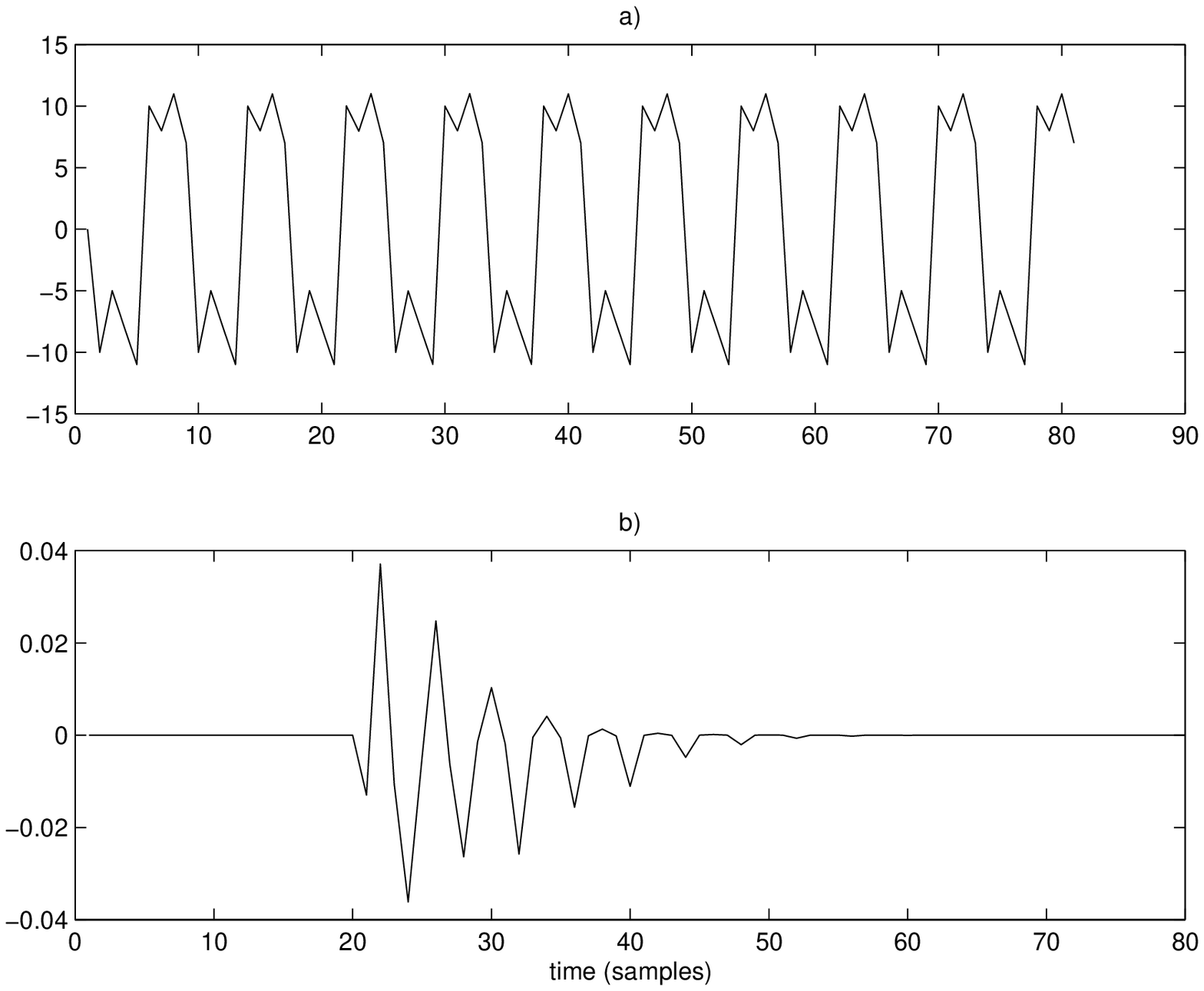,width=9.0cm}
  \caption{Closed loop system: rejection of additive noise. a) Time evolution of the system.
  b) Distance from the target evolution when noise is added to the loop,
  from sample 21 to sample 25}
  \label{halfpersimu}
\end{figure}
\fi

One might be curious about the possibility of reaching a desired
stable periodic motion from a quasi-zero random state, controlling
the slope in the origin as in the previous example. Despite the
fact that the solution appears to be in the combined use of the
skills given in the previous examples, whether such control would
be possible or not with a time-invariant 1-in 1-out non-linear
map, seems to be a non-trivial problem.

If the filter $H(z)$ is not omitted in $G(z)$, the control of
stability of the single loop feedback system of Fig.
\ref{feedback} can be conveniently approached by studying the
Jacobian matrix $J$ of the map $F$ which describes the state
transition ${\mathbf x}(n+1)=F({\mathbf x}(n))$ at every
successive time step, $\mathbf{x}$ being the global state of the
system. Let the linear element $G(z)$ be, as before, the cascade
of a delay line $D_L(z)=z^{-L}$ of length $L$ and a low-pass
filter $H(z)$. We are interested in leading the system to a stable
periodic motion. In a steady state situation, the state ${\mathbf
x}_L=[x_1,x_2,\dots,x_L]^T$ of the delay line undergoes a linear
distortion due to the filtering stage. This is represented by the $L$-point circular discrete-time convolution~\cite{OppenheimAndSchafer} 
\begin{equation}
\tilde{y}_k = (h \circledast x)_k, 1\leq k\leq L \; .
\end{equation}
To restore the original state of the delay line
the non-linear map $f$ can be shaped on the base of a training set
given by equation~(\ref{eqn:cup}), with
\begin{equation}
{\mathcal Y}_1=\bigcup_{i=1}^{L}(\tilde{y}_i,y_{L+i}),
\end{equation}
and
\begin{equation}
{\mathcal Y}_2=\bigcup_{i=1}^{L}(\tilde{y}_{L+i},y_i).
\end{equation}

%\begin{equation}
%{\mathcal Y}={\mathcal Y}_1 \cup {\mathcal Y}_2= \left\{
%(\tilde{y}_{i},y_{L+i}) \right\} \cup \left\{
%    (\tilde{y}_{L+i},y_i)
%\right\}.
%\end{equation}

In general, the geometric locus given by the training set
${\mathcal Y}$ will not necessarily be a curve of dimension 1, and
the map will need to be unfolded in a higher dimensional space. We
consider here the case where a one-dimensional map is sufficient
to our purposes. If $H(z)$ is a first order FIR filter with
coefficients $b_1$ and $b_2$, the system can be given in its state
space form as
\begin{equation}
{\mathbf x}(n+1) = \left[\begin{array}{ccccc}
     0 & &\cdots & & 0 \\
     b_1 & 0 & & & 0 \\
     0 & 1 & 0 & & 0 \\
     \vdots & & \ddots &\ddots & \vdots \\
     0 & & \cdots & 1 & 0
  \end{array}\right]
  {\mathbf x}(n)+\left[\begin{array}{c}
   f(x_L) \\
    b_2f(x_L) \\
    0 \\
    \vdots\\
    0
  \end{array}\right],
\end{equation}
where ${\mathbf x}(n) = [x_0(n) x_1(n)\cdots x_L(n)]^T$ is the
global state of the system at time $n$.

The Jacobian matrix of the state transition map, evaluated in
$\hat{{\mathbf x}}=[\hat{x}_0 \hat{x}_1 \cdots \hat{x}_L]^T$, is
given by
\begin{equation}
{\mathbf J}(\hat{{\mathbf x}})=\left[\begin{array}{ccccc}
     %0 & & \cdots & & \frac{\partial f}{\partial x_L}|_{x_L=\hat{x}_L}\\
     %b_1 & 0 & & & b_2 \frac{\partial f}{\partial x_L}|_{x_L=\hat{x}_L} \\
     0 & & \cdots & & d\\
     b_1 & 0 & & & b_2d\\
     0 & 1 & 0 & & 0 \\
     \vdots & &\ddots & \ddots & \vdots  \\
     0 & & \cdots & 1 & 0
  \end{array}\right],
\end{equation}
where
\begin{equation}
d\triangleq\frac{\partial f}{\partial x_L}|_{x_L=\hat{x}_L} \; .
\end{equation}
A periodic orbit $[\hat{{\mathbf x}}(n) \cdots \hat{{\mathbf
x}}(n+2L-1)]$ of period $2L$ is asymptotically stable if the
Jacobian ${\mathbf J}$ has eigenvalues of magnitude less than one
for each point of the periodic orbit.
%if the Jacobian ${\mathbf M}$ of the $2L$-times iterated map,
%i.e. ${\mathbf M}={\mathbf J}^{2L}={\mathbf J}(\hat{{\mathbf
%x}}(n))\cdot {\mathbf J}(\hat{{\mathbf x}}(n+1)) \cdots {\mathbf
%J}(\hat{{\mathbf x}}(n+2L-1))$, has eigenvalues of magnitude less
%than one.
The eigenvalues of ${\mathbf J}$ are the roots of the polynomial
$z^{L+1}-b_1dz-b_2d$, and are plotted in Fig. \ref{jeigenvalues}
for $L=3$, and for different values of $d$. The lower and the
upper figures refer to two different low-pass filters $H(z)$.

\if F\draft
\begin{figure}[h]
  \centering
%\psfull
\psfig{file=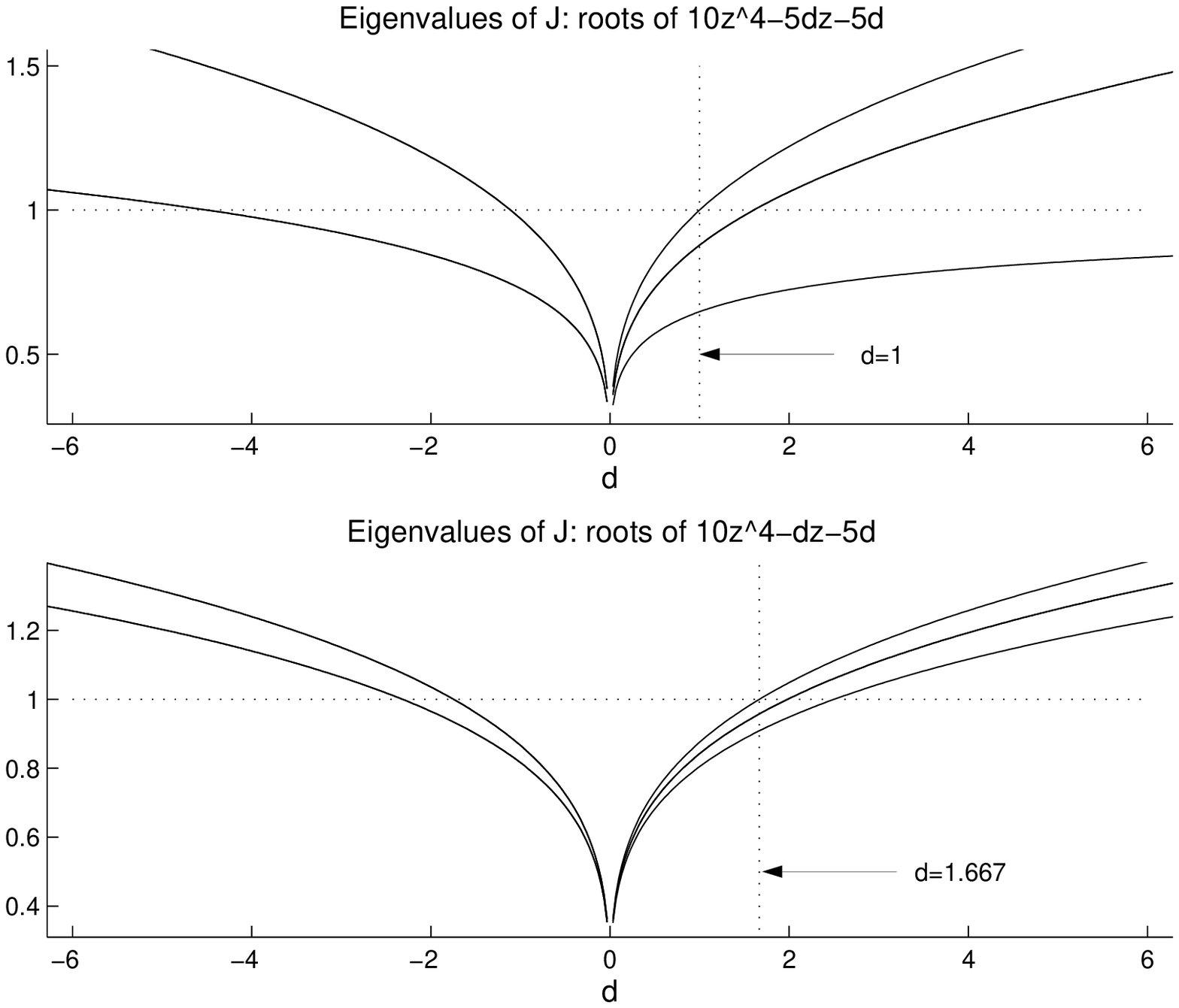,width=9.0cm}
  \caption{Magnitude of the roots of the polynomial $z^4-b_1dz-b_2d$ (or eigenvalues of the Jacobian
   matrix $J$) for $b_1=b_2=0.5$ (upper figure) and $b_1=0.1$, $b_2=0.5$ (lower figure).}\label{jeigenvalues}
\end{figure}
\fi
Let us focus the attention on the case where $L=3$ and $H(z)$ has
coefficients $b_1=0.1$ and $b_2=0.5$. From Fig. \ref{jeigenvalues}
it can be seen that ${\mathbf J}$ has eigenvalues $|\lambda|<1$ if
$|d|<1.667$. Thus, in order to have a stable and noise-robust
periodic solution, the magnitude of the derivative of the map in
each point of the training data must not exceed $1.667$.

Usually, a perturbation to the closed loop system is modeled with
random noise added to the loop at a given point. However, it can
be of some interest to vary the parameters of the linear
components in the loop such as, for example, the low-pass filter
$H(z)$. This can be useful to control the spectral content of the
resulting time series. The stability of the whole system is thus
investigated by applying, for a short time window, a perturbation
to the filter coefficients $b_1$ and $b_2$. Fig. \ref{3errors}
shows the reaction of the system to a random perturbation, with
an upper bound in magnitude of $0.02$, occurring at sample time
$44$ and ending at sample time $48$. It can be seen that the
perturbation will be persistent for values of $|d|$ higher than
$1.667$ (upper figure), and that it will be rejected for
$|d|<1.667$ in a time that is shorter the lower we choose $|d|$
(middle and lower figures).

%\begin{figure}[h]
%  \centering
%  \includegraphics[scale=.5]{mapident.eps}
%  \caption{}\label{mapident}
%\end{figure}
\if F\draft
\begin{figure}[h]
  \centering
%\psfull
\psfig{file=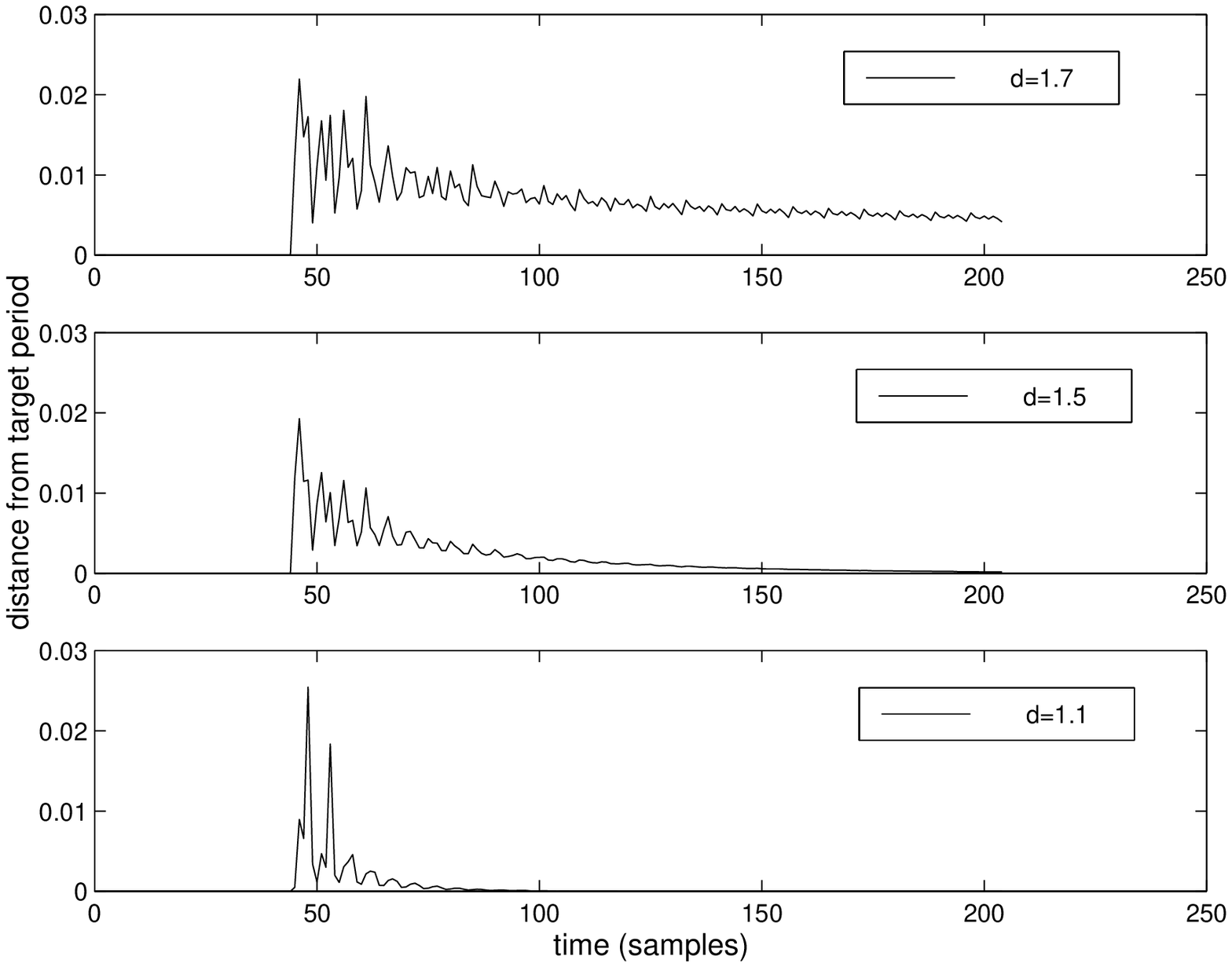,width=9.0cm}
  \caption{Rejection of a perturbation of the coefficients of the filter $H(z)$. The
  perturbation $b_1+\Delta b_1$, $b_2+\Delta b_2$, occurring at sample time $44$ and
  ending at sample time $48$, had upper bound in magnitude $|\Delta b_i|\leq 0.02$, $i=1,2$.
  }\label{3errors}
\end{figure}
\fi
\section{Conclusions}\label{sec:concl}
The use of the Orthogonal Least Squares algorithm to approximate a
non-linear map with arbitrary derivatives with radial basis
function networks has been investigated. A modified version of the
classic OLS algorithm formulation has been proposed, which uses
the same orthogonalization approach for both the regressors of the
map and the regressors of its derivatives. The usefulness of the
method has been illustrated on application examples from the field of
control of single loop feedback systems, and we have stressed the importance of
 derivatives of the non-linear map  to control important
features such as stability, velocity of transients, and rejection
of disturbances.

\bibliographystyle{IEEEbib}
\bibliography{dsp_nlchaos,dsp_instr}

\if T\draft
\clearpage
\section*{Illustrations}
\clearpage
\begin{figure}[ht]
  \begin{center}
\psfull
\psfig{file=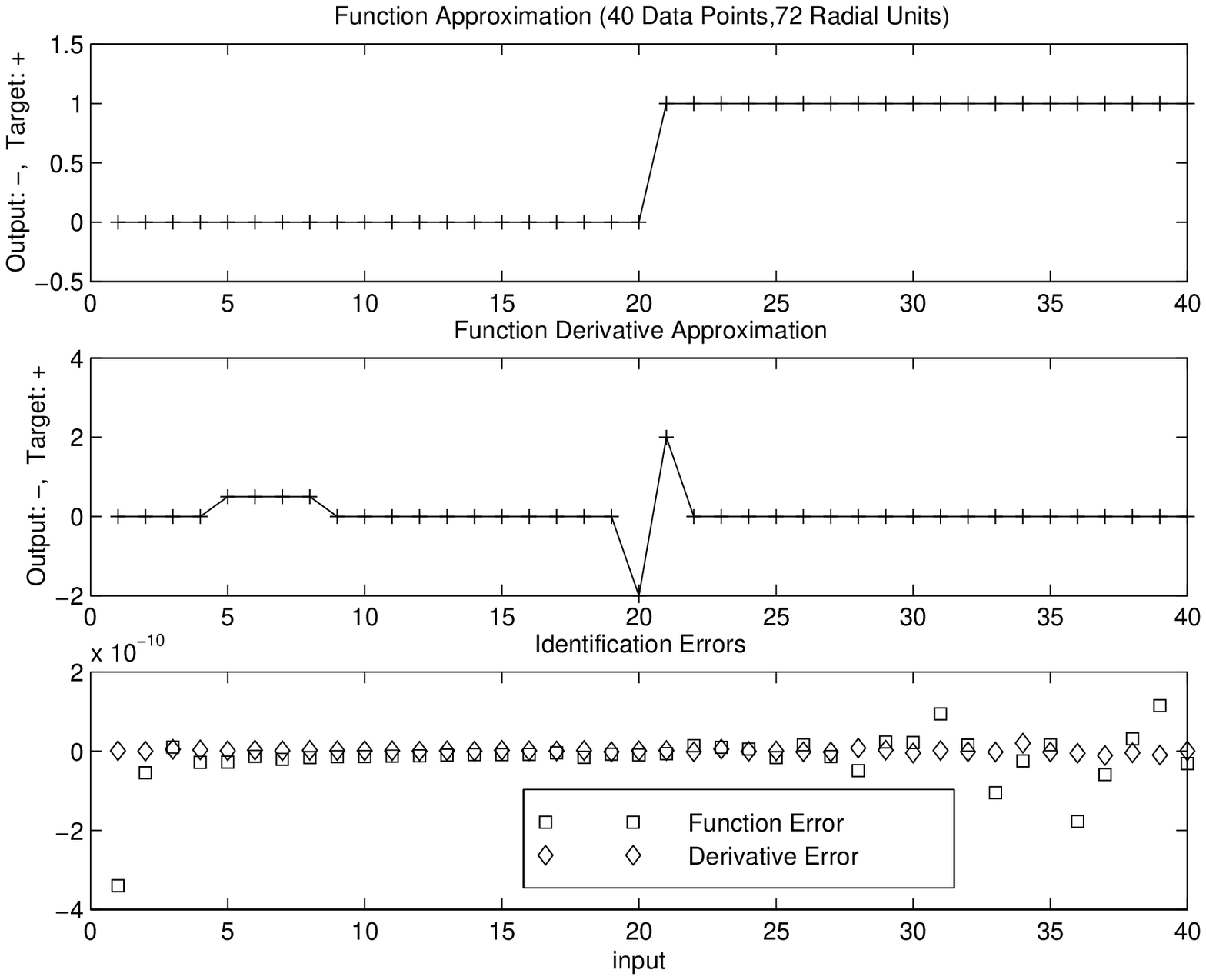}
\caption{Result of the training
  procedure applied to the fitting of a step-like data set ($+$, upper figure),
  with arbitrary derivative constraint ($+$, middle figure). In upper and middle figures,
  $+$ is the desired output and the continuous line is the actual output.
  The problem required $72$ radial units to fit $40$ data points, with an
  identification error less than $10^{-9}$ in magnitude.}\label{step1}
  \end{center}
\end{figure}

\clearpage

\begin{figure}[ht]
  \centering
\psfull
\psfig{file=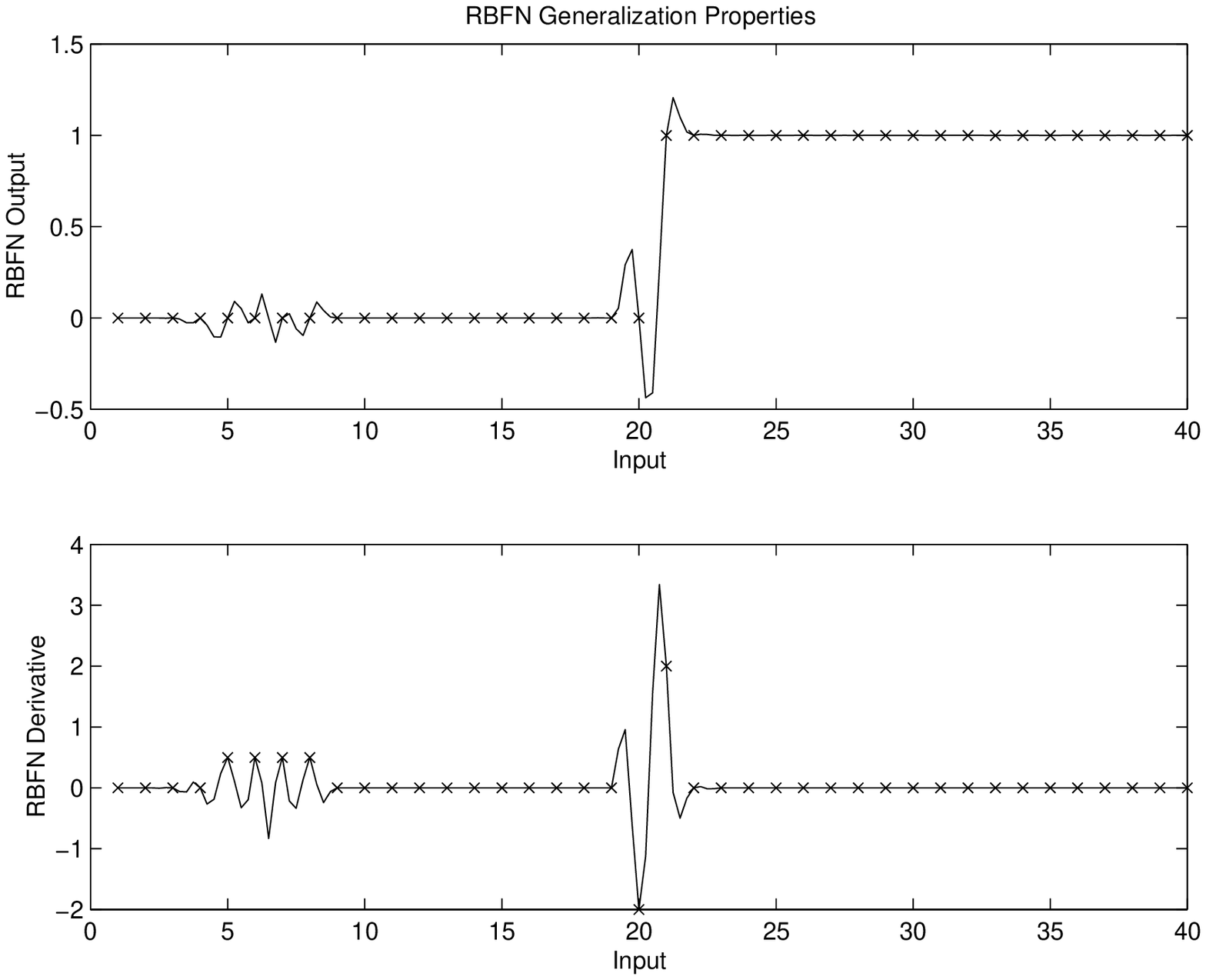}
  \caption{Interpolation properties of the
  identified RBF Network: the output of the model was computed on a input set which is
  denser than the original one (x: data set).}\label{step2}
\end{figure}

\clearpage

\begin{figure}[ht]
  \centering
\psfull
\psfig{file=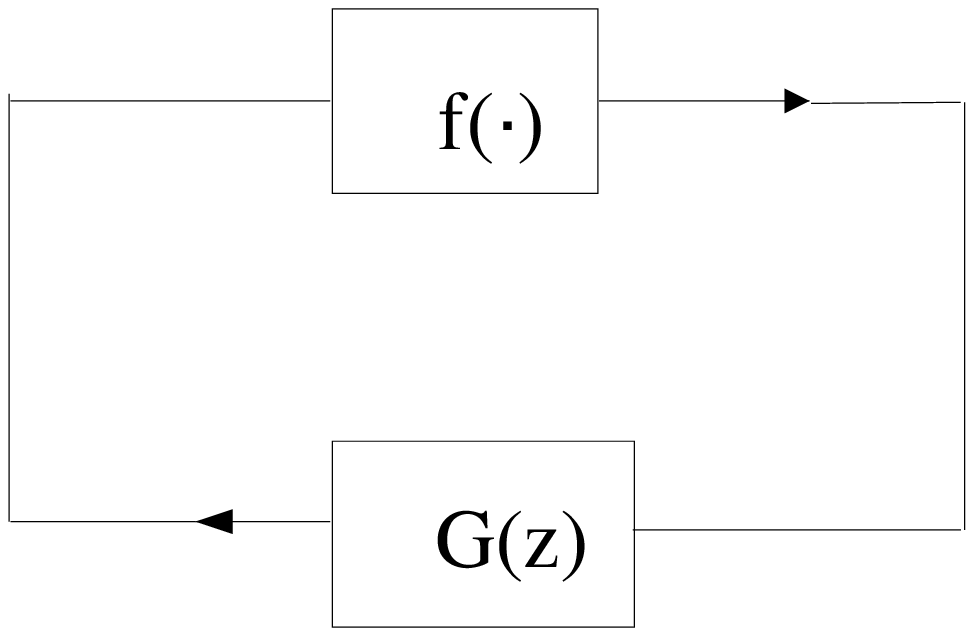}
\caption{Single loop feedback system}
\label{feedback}
\end{figure}

\clearpage

\begin{figure}[ht]
  \centering
\psfull
\psfig{file=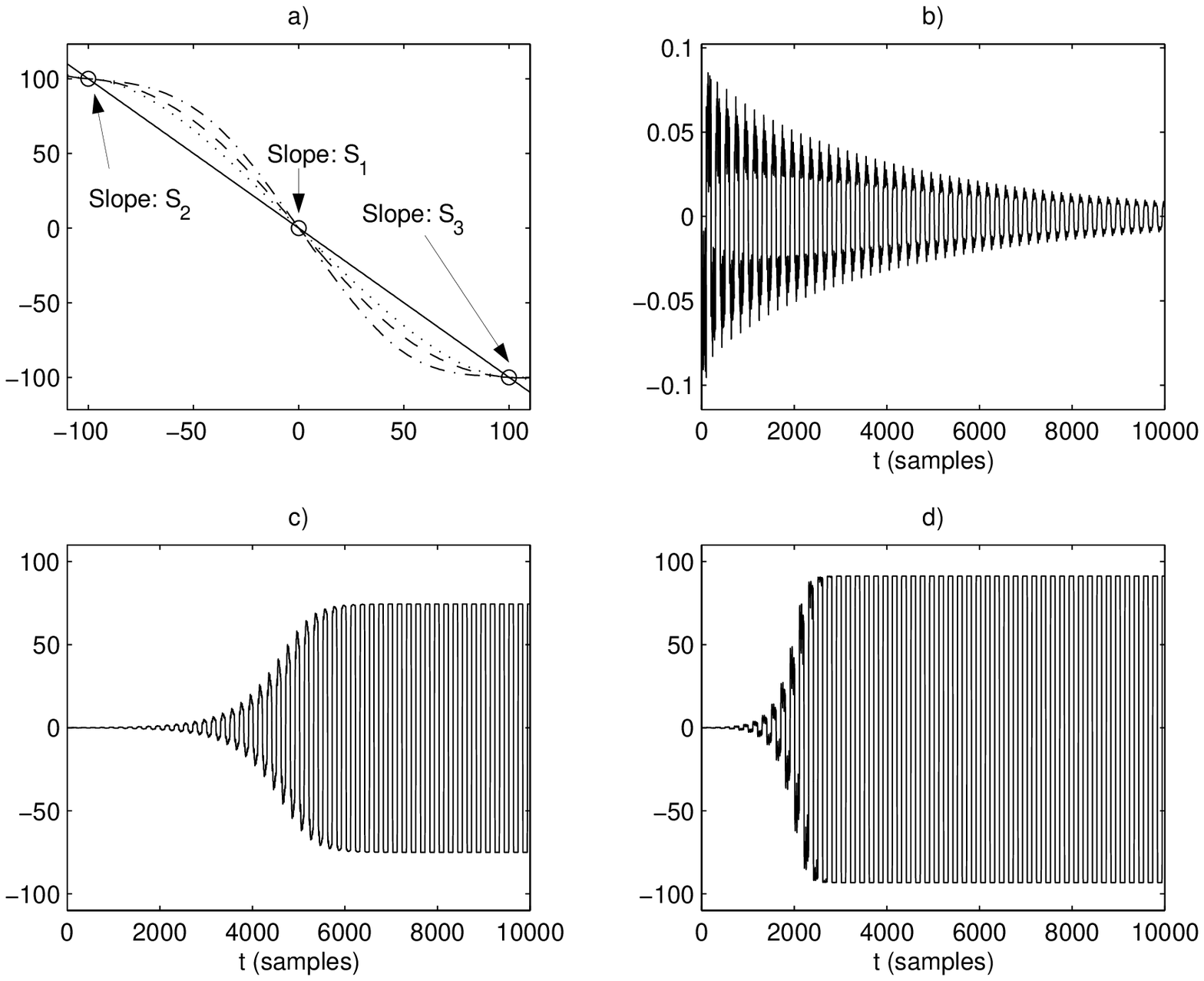}
\caption{Simulation of the circuit for different
  values of the derivative of the fixed point in the origin. a) shapes of the function $f(\cdot)$ for
  $S_1=-1.4$ (dotted line),  $S_1=-1.65$ (dashed line) and  $S_1=-2$ (dashdotted
  line). b,c,d) time evolution from random initial conditions in the three cases. The slope
  $S_2=-0.1$ of the other two fixed points is held constant in the three cases,
  and limits the growth of the system evolution.}\label{chuatest1}
\end{figure}

\clearpage

\begin{figure}[ht]
  \centering
\psfull
\psfig{file=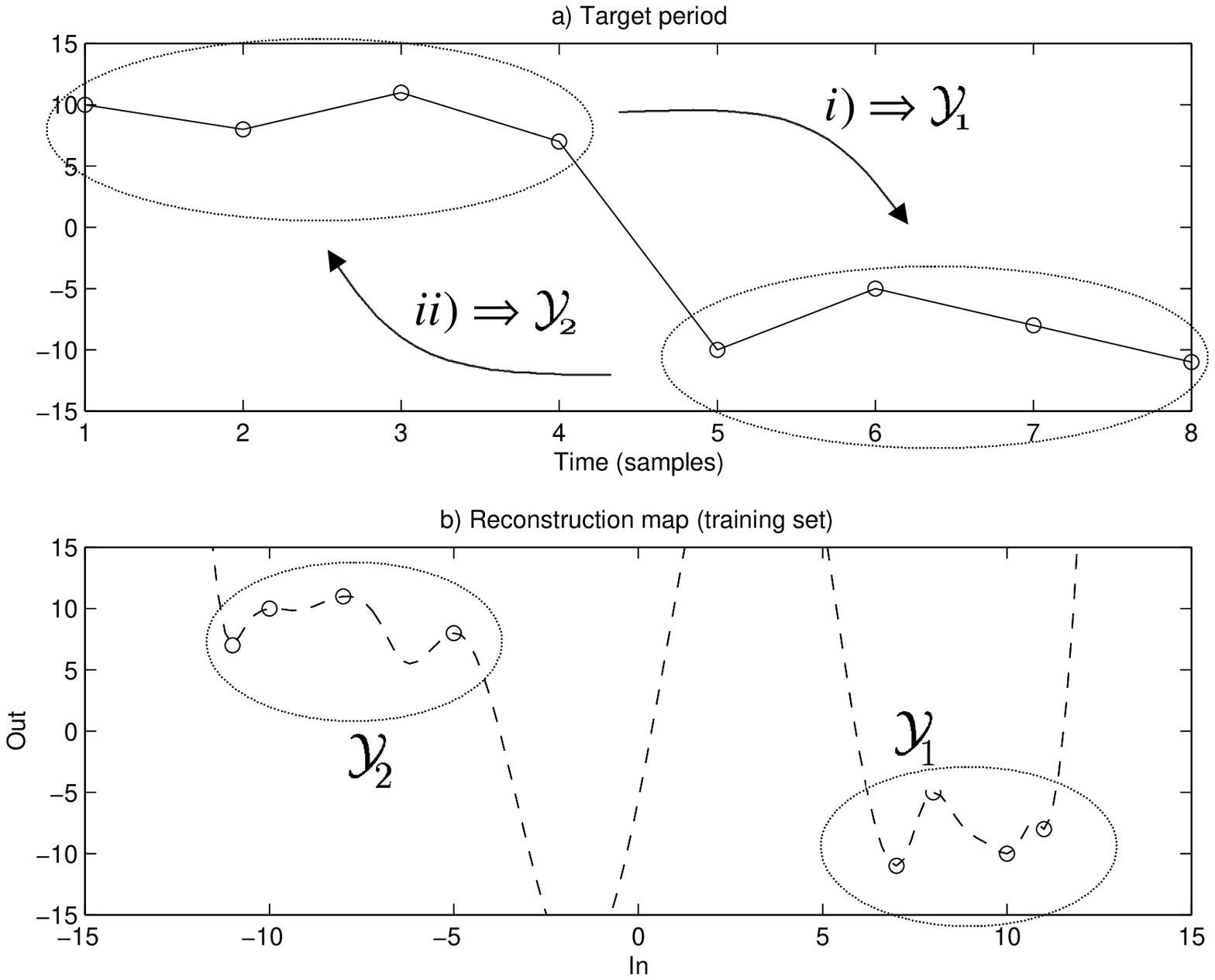}
  \caption{Training data (a) and approximation of the unknown
  function $f$ (b, dashed curve). A desired derivative of $0.3$ in
magnitude was imposed for all eight data points}
  \label{halfper2id}
\end{figure}

\clearpage

\begin{figure}[ht]
  \centering
\psfull
\psfig{file=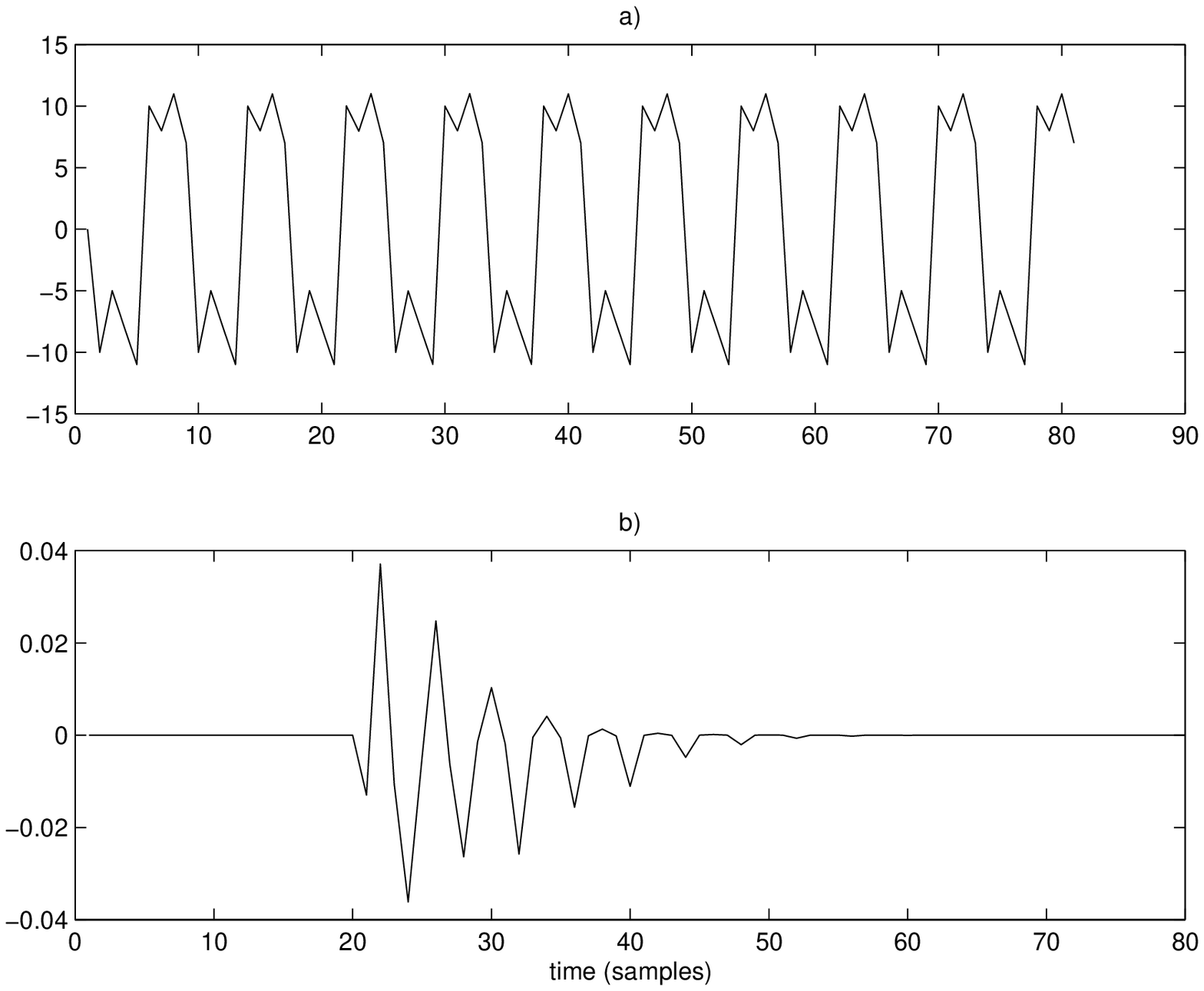}
  \caption{Closed loop system: rejection of additive noise. a) Time evolution of the system.
  b) Distance from the target evolution when noise is added to the loop,
  from sample 21 to sample 25}
  \label{halfpersimu}
\end{figure}

\clearpage

\begin{figure}[ht]
  \centering
\psfull
\psfig{file=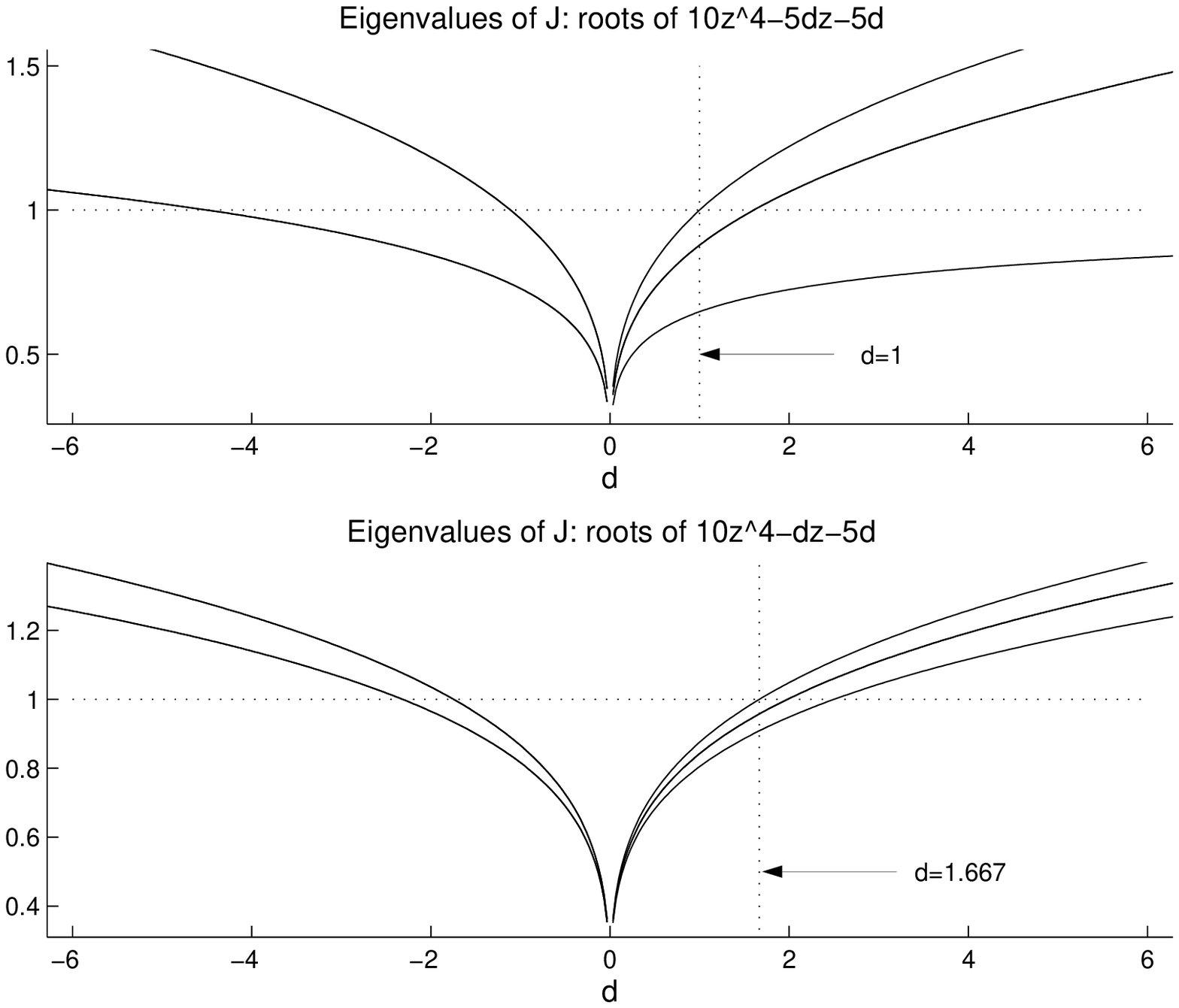}
  \caption{Magnitude of the roots of the polynomial $z^4-b_1dz-b_2d$ (or eigenvalues of the Jacobian
   matrix $J$) for $b_1=b_2=0.5$ (upper figure) and $b_1=0.1$, $b_2=0.5$ (lower figure).}\label{jeigenvalues}
\end{figure}

\clearpage

\begin{figure}[ht]
  \centering
\psfull
\psfig{file=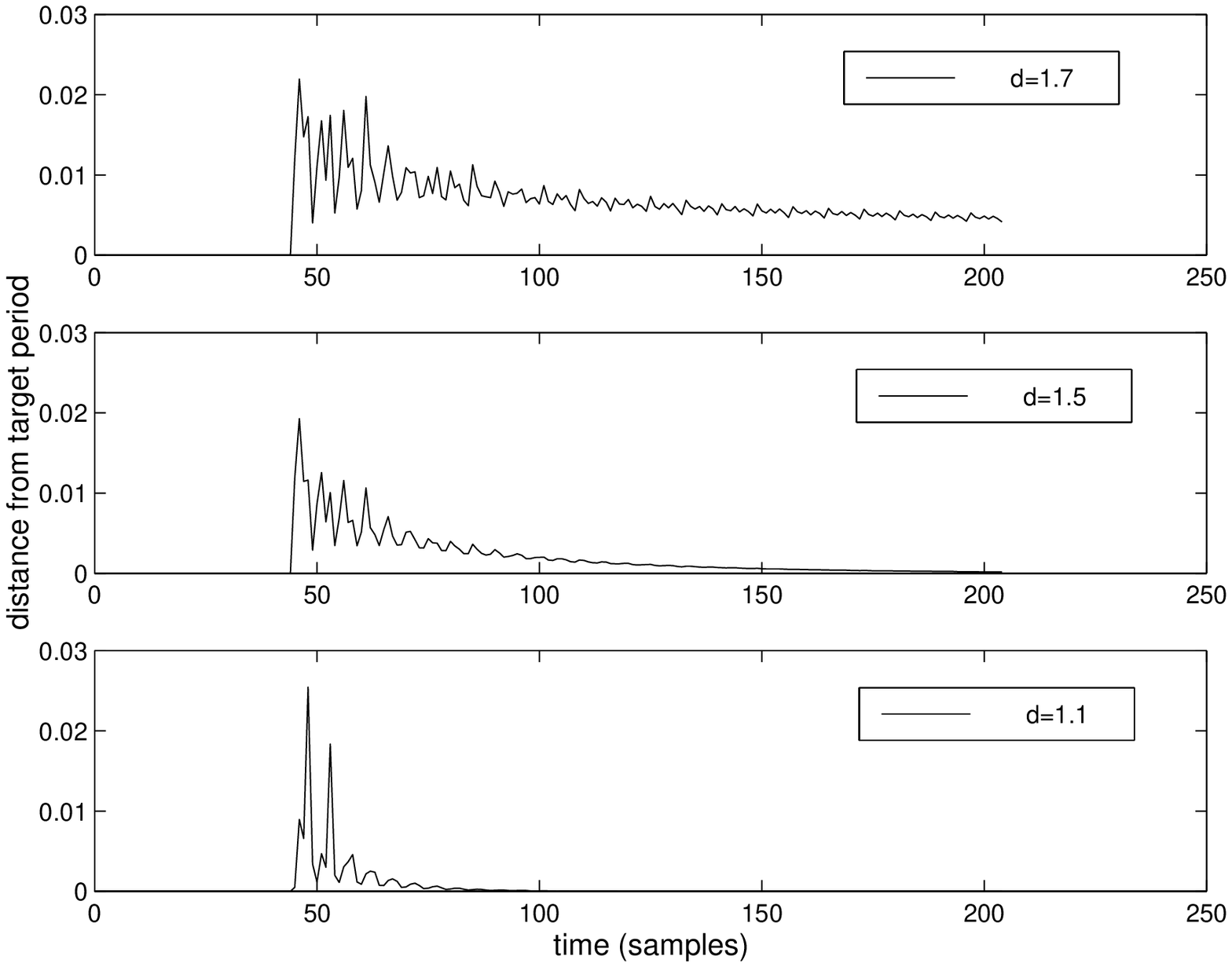}
  \caption{Rejection of a perturbation of the coefficients of the filter $H(z)$. The
  perturbation $b_1+\Delta b_1$, $b_2+\Delta b_2$, occurring at sample time $44$ and
  ending at sample time $48$, had upper bound in magnitude $|\Delta b_i|\leq 0.02$, $i=1,2$.
  }\label{3errors}
\end{figure}

\clearpage

\listoffigures

\clearpage

\section*{List of Footnotes}

\begin{itemize}

\item Manuscript received \ldots 

\item Carlo Drioli is with the Dipartimento di Elettronica e Informatica,
Universit\`a di Padova, 35131 Padova, Italy (e-mail: adrian@dei.unipd.it) 

\item Davide Rocchesso is with the Dipartimento Scientifico e Tecnologico,
Universit\`a di Verona, 37134 Verona, Italy (e-mail: rocchesso@sci.univr.it) 

\item[1] {Note that in this case the length of vector ${\mathbf
w}$ and the number of columns of matrices ${\mathbf P}$ in equation
(\ref{vecols2}) is $h$ instead of $H$}

\end{itemize}

\fi

\end{document}